\documentclass[conf]{new-aiaa}
\usepackage[utf8]{inputenc}

\usepackage{graphicx}
\usepackage{amsmath}
\usepackage[version=4]{mhchem}
\usepackage{siunitx}
\usepackage{longtable,tabularx}
\usepackage{custom}
\usepackage{thmtools}
\usepackage{cleveref}
\usepackage{booktabs} 
\newtheorem{theorem}{Theorem}[section]

\newtheorem{definition}[theorem]{Definition}

\title{Physics-Informed Machine Learning  \\ for Characterizing System Stability}

\author{Tomoki Koike\footnote{PhD Candidate, School of Aerospace Engineering,
AIAA Student Member, tkoike@gatech.edu} and Elizabeth Qian\footnote{Assistant
Professor, Schools of Aerospace Engineering \& Computational Science and
Engineering, AIAA Member, eqian@gatech.edu}} \affil{Georgia Institute of
Technology, Atlanta, Georgia, 30332}

\begin{document}

\maketitle

\begin{abstract}

    In the design and operation of complex dynamical systems, it is essential to ensure that all state trajectories of the dynamical system converge to a desired equilibrium within a guaranteed stability region. Yet, for many practical systems---especially in aerospace---this region cannot be determined a priori and is often challenging to compute. One of the most common methods for computing the stability region is to identify a Lyapunov function.  A Lyapunov function is a positive function whose time derivative along system trajectories is non-positive, which provides a sufficient condition for stability and characterizes an estimated stability region.  However, existing methods of characterizing a stability region via a Lyapunov function often rely on explicit knowledge of the system governing equations. In this work, we present a new physics-informed machine learning method of characterizing an estimated stability region by inferring a Lyapunov function from system trajectory data that treats the dynamical system as a black box and does not require explicit knowledge of the system governing equations. In our presented Lyapunov function Inference method (LyapInf), we propose a quadratic form for the unknown Lyapunov function and fit the unknown quadratic operator to system trajectory data by minimizing the average residual of the Zubov equation, a first-order partial differential equation whose solution yields a Lyapunov function. The inferred quadratic Lyapunov function can then characterize an ellipsoidal estimate of the stability region. Numerical results on benchmark examples demonstrate that our physics-informed stability analysis method successfully characterizes a near-maximal ellipsoid of the system stability region associated with the inferred Lyapunov function without requiring knowledge of the system governing equations. 
\end{abstract}

\section{Nomenclature}

{\renewcommand\arraystretch{1.06}
\noindent\begin{longtable*}{@{}l@{\quad=\quad}l@{}}
\( t \) & Time variable. \\
\( n \) & State dimension. \\
\( \R^n, \R^{m\times n} \) & Space of real \(n\) vectors and \( m\times n \) matrices. \\
\( \xvec \) & State vector, \( \xvec(t)\in\R^n \). \\
\( \dot{\xvec} \) & Time derivative, \( \dot{\xvec} = \tfrac{d\xvec}{dt}\in\R^n \). \\
\( f(\xvec) \) & Nonlinear function, \( f:\R^n\to\R^n \), defining \( \dot{\xvec} = f(\xvec) \). \\
\( \mathcal{D}_0 \) & True stability region. \\
\( \mathcal{D}(c) \) & Estimated stability region: \( \{\xvec : V(\xvec)\le c, \dot{V} < 0\} \). \\
\( V,\widetilde{V} \) & True and approximate Lyapunov functions, \( V,\widetilde{V}:\R^n\to\R \). \\
\( h(\xvec) \) & Auxiliary function in Zubov equation, \( h:\R^n\to\R \). \\
\( \Pmat \) & Matrix parametrizing the quadratic Lyapunov function \( V(\xvec)=\xvec^\top\Pmat\xvec \). \\
\( \Qmat \) & Matrix parametrizing the quadratic form \( h(\xvec)=\xvec^\top\Qmat\xvec \). \\
\( N \) & Number of snapshot data points. \\
\( \Xmat=[\xvec(t_1),\dots,\xvec(t_N)] \) & State snapshot matrix in \( \R^{n\times N} \). \\
\( \dot{\Xmat}=[\dot{\xvec}(t_1),\dots,\dot{\xvec}(t_N)] \) & Derivative snapshot matrix in \( \R^{n\times N} \). \\
\( \otimes \) & Kronecker product. \\
\( \|\cdot \|_2,\|\cdot \|_F \) & Euclidean and Frobenius norms. \\
\( \mathcal{X} \) & Specified operating region of a system.
\end{longtable*}}
\section{Introduction}\label{sec:intro}
In the design of complex nonlinear systems, engineers encounter phenomena such as bifurcations, limit cycles, and chaotic attractors. Stable limit cycles can support sustained periodic operation, whereas instabilities can drive trajectories away from equilibrium into unsafe regimes. To distinguish safe from unsafe initial conditions of such different system behaviors, engineers can study the \emph{stability region} (also called the region of attraction or domain of attraction), defined as the set of initial states whose trajectories converge to the equilibrium. Knowing this region reveals the safe operating regime and how far nonlinear effects may be pushed before stability is lost. Reliable estimation of the stability region can therefore provide insight into the system's response to varying initial conditions, perturbations, and parameters, guiding decision-making and design for systems in various fields, including robotics~\cite{dawson2023safe,brunke2022Safe}, power systems~\cite{zhang2020review}, and fluid dynamics~\cite{schmid2014Analysis}. In aerospace applications, stability region estimations led to enlarging safe flight envelopes for an F-8 aircraft short-period model~\cite{biswas2024design} and identifying the safe regions of a 4D longitudinal dynamics of NASA's Generic Transport Model (GTM) aircraft~\cite{lai2021Nonlinear}.

One of the most common methods for assessing system stability is Lyapunov's method. This method assesses stability by finding a smooth, \emph{energy-like} function \( V(\xvec) \) defined to be zero at the equilibrium and strictly positive for all other states in a specified region. Additionally, \( V \) must always be non-increasing along every trajectory in that same region, i.e., its time derivative is non-positive. If those two conditions hold everywhere in the state space, the equilibrium is globally asymptotically stable; if they hold only inside some region around the equilibrium (the stability region), they guarantee only local stability. Moreover, the sublevel sets of \(\ V \) serve as inner estimates of the stability region~\cite{lyapunov1992}. Thus, identifying a Lyapunov function is a key step in designing and validating engineering systems that must remain stable despite inherent variations and uncertainties.

Most classical Lyapunov-based methods rely on explicit knowledge of the system's governing equations to determine a stability region by constructing a Lyapunov function. Analytical methods may assume a parametric form for the gradient of \(V\) and solve for coefficients that enforce \(\dot V<0\) (variable-gradient method~\cite{schultzVariableGradientMethod1962}), or solve the Zubov equation---a first order partial differential equation (PDE) which is known to govern the dynamics of the Lyapunov equation---via polynomial or rational expansions~\cite{zubov1964,margolis1963,ferguson1970}. Advances in convex optimization have given rise to computational approaches that pose stability certification as linear programs for piecewise-linear \(V\)~\cite{braytonStabilityDynamicalSystems1979,pedroPiecewiself1999} and semidefinite programs (SDPs) for quadratic or higher-order Lyapunov functions~\cite{boyd1994linear,JohanssonCPQ1998}. These SDPs include classical Krasovskii's matrix-inequality formulation~\cite{krasovskii1963problems} and modern sum-of-squares (SOS) programs, which enforce nonnegativity of \(V\) by requiring a polynomial to decompose into sums of squares~\cite{parrilo2000structured,papachristodoulouCLF2002}. We emphasize that all of these methods require knowledge of the governing equations \emph{a priori} to set up either the analytical conditions or the constraints that guarantee \(\dot V<0\), thereby certifying the equilibrium's stability region. However, many complex multidisciplinary systems such as those in aerospace design lack fully known governing equations, precluding direct application of these tools and motivating data-driven stability analysis methods.

Building on this motivation, data-driven approaches for estimating safe stability regions have gained increasing attention~\cite{drgona2025Safe}.
For example,~\cite{barreau2024Learninga} proposes a Lyapunov function in Taylor-series form: the quadratic term is obtained via an SDP-based linear matrix inequality, and higher-order terms are modeled with a neural network. Although their approach requires no simulated trajectory data, it still relies on explicit knowledge of the system's governing equations, since the Taylor-neural Lyapunov function is trained under a reformulated Zubov-equation constraint that directly maximizes the stability region. Another line of data-driven methods leverages the Koopman operator, which is an infinite-dimensional linear operator that evolves observables (functions of the system states) over time. In practice, a finite set of the Koopman operator's leading eigenfunctions is learned, and then the Lyapunov function is constructed by forming a linear combination of those spectral components~\cite{mauroy2013spectral,mauroy2016Global}. In~\cite{umathe2024Spectral}, the authors further demonstrate how such eigenfunctions can be used to approximate the boundary of the stability region directly using the eigenvalues and eigenfunctions of the Koopman operator. Another work~\cite{deka2022Koopmanbased} partially automates the selection of observables by learning the eigenfunctions using autoencoder networks. However, when the system governing equations are unknown, selecting an appropriate dictionary of observables or tuning the autoencoder to ensure the observables span the key modes of the infinite-dimensional operator remains a challenge.

In this work, we introduce a scientific machine learning approach that embeds domain knowledge, in the form of the Zubov equation, directly into a data-driven Lyapunov function inference method. We restrict \(V(\xvec)\) to a quadratic form and learn its quadratic operator by minimizing the average residual of the Zubov equation over available snapshot and time derivative data. This physics-informed training enforces the \( \dot{V} < 0 \) condition and produces an ellipsoidal estimate of the stability region from trajectory data, without requiring explicit knowledge of the system's governing equations. 
Another physics-informed machine learning work using data to approximate solutions to the Zubov equation is~\cite{kang2024data}, which trains a neural network on simulation outputs of an augmented dynamical system to approximate an integral-form solution of the Zubov equation. In contrast, our approach fits our candidate Lyapunov function directly to trajectory data from the system itself.
Our approach is related to that of~\cite{liu2025physics,zhou2024PhysicsInformed}, which fits a neural network form Lyapunov function to system trajectory data by minimizing the residual of the Zubov equation. However, while neural networks offer considerable flexibility in expressing complex nonlinear functions, they require large amounts of data to train and are difficult to interpret: verification of the associated domain of attraction in~\cite{liu2025physics,zhou2024PhysicsInformed} requires the use of satisfiability modulo theory (SMT) solvers to verify formal conditions relating the domain of attraction of the neural network to that of an associated quadratic Lyapunov function. In contrast, our direct consideration of a quadratic form requires less data to train and allows the resulting Lyapunov function to be more easily interpreted and analyzed.
There also exist works which combine Zubov's theory with Koopman operators:~\cite{meng2023Learning} solves a modified Zubov equation in the lifted Koopman space, and~\cite{zhou2025Learning} extends this framework to jointly learn both the governing dynamics and the Lyapunov function as well as the stability region. 
Although these Zubov-Koopman methods exploit the Koopman operator's ability to represent complex nonlinear dynamics as linear evolution in an infinite-dimensional function space and thus capture diverse behaviors via a rich dictionary of observables, they still rely on a neural network embedded in the framework to approximate the Zubov equation solution. This reliance restricts them to an SMT-solver-based verification procedure (as in~\cite{liu2025physics}) and makes their internal structure challenging to interpret compared to our explicit quadratic Lyapunov function.

The contributions of this article are: (i) a new physics-informed method of learning a stability region by inferring a quadratic Lyapunov function from system trajectory data that treats the dynamical system as a black box and does not require explicit knowledge of the system governing equations; and (ii) numerical demonstration on benchmark dynamical systems that shows how our approach successfully identifies a near-maximal ellipsoid estimate of the stability region for a given operating regime and its corresponding Lyapunov function. 

The remainder of this work is organized as follows. \Cref{sec:2-background} describes the problem formulation and the key theories by Lyapunov and Zubov. \Cref{sec:3-lyapinf} introduces our method. \Cref{sec:4-examples} demonstrates our method with several numerical examples. Finally, \Cref{sec:conclusion} concludes and discusses directions for future work.

\section{Background}\label{sec:2-background}

\subsection{System Definition}\label{sec:2.1-system-definition}
Consider an autonomous \(n\)-dimensional nonlinear dynamical system
\begin{equation}\label{eqn:nonlinear-sys}
\dot{\xvec}(t) = f(\xvec(t)), \quad \xvec_0 = \xvec(0), \quad t \in [0, t_f],
\end{equation}
where \(\xvec(t) \in \R^n\) denotes the state vector of the system, with \(t\) in a given interval \([0, t_f]\) where \(0 < t_f < \infty \). The function \(f: \R^n \to \R^n\) is assumed to be locally Lipschitz continuous to ensure the existence and uniqueness of a solution to the initial value problem~\eqref{eqn:nonlinear-sys}. Although we express the dynamics explicitly in terms of \(f\), our method treats \(f\) as a black box and does not require knowledge of its form.

\subsection{Lyapunov Stability Theory}\label{sec:2.2-lyapunov-stability-theory}
\begin{figure}[htb!]
    \centering
    \includegraphics[width=0.65\linewidth]{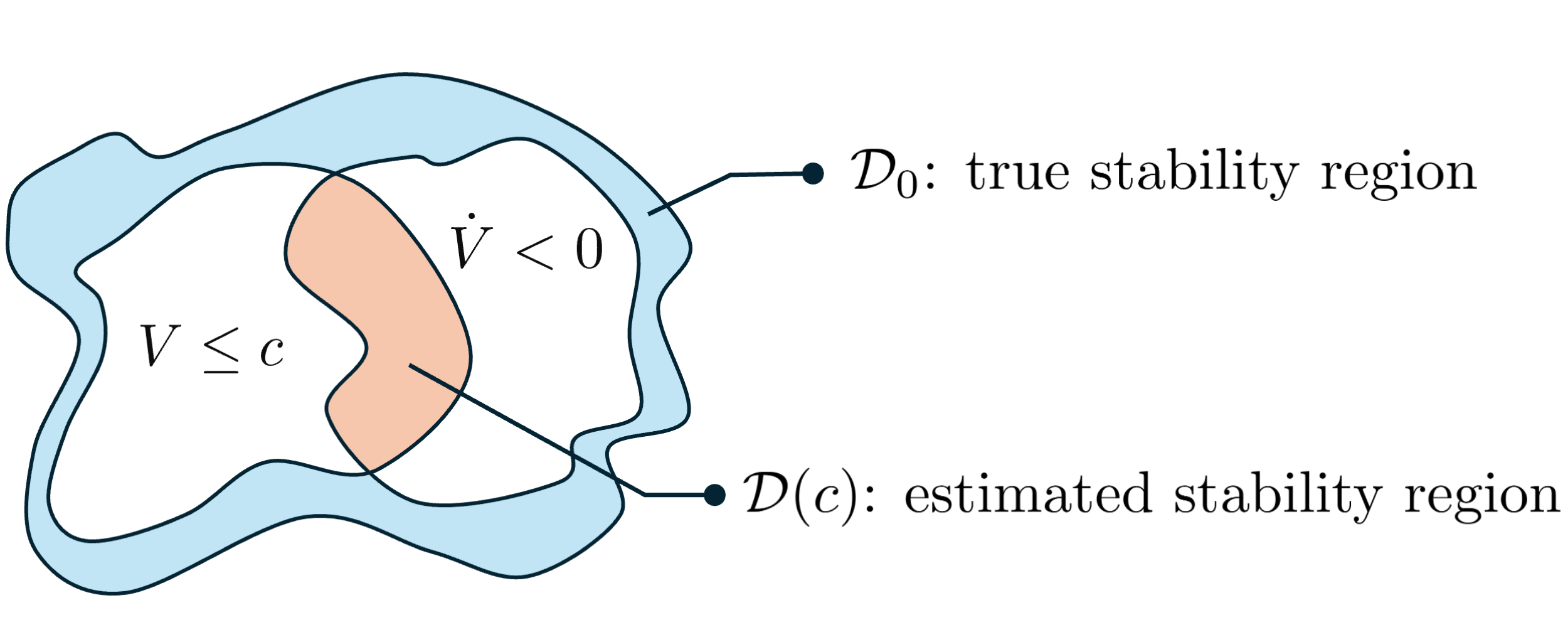}
    \caption{Illustration of the estimated stability region characterized by some Lyapunov function \(V\).}\label{fig:doa-diagram}
\end{figure}

Without loss of generality, let the origin be an equilibrium point of~\eqref{eqn:nonlinear-sys}. The stability of the system about the origin is defined as follows~\cite{Haddad2008-hg}:
\begin{definition}\label{def:stability}
Consider the nonlinear dynamical system (\ref{eqn:nonlinear-sys}). The origin is said to be \textit{Lyapunov stable} if, for any \( \epsilon > 0 \) there exists a \( \delta > 0 \), such that \( \|\mathbf{x}(t)\| < \epsilon \) is true for all \( t \geq 0 \) whenever \( \|\mathbf{x}_0\| < \delta \). Moreover, the origin is \textit{asymptotically stable} if it is Lyapunov stable and there exists a \( \delta > 0 \) such that \( \lim_{t\to\infty}\mathbf{x}(t) = 0 \) whenever \( \|\mathbf{x}_0\| < \delta \).
\end{definition}
Lyapunov stability merely guarantees boundedness of trajectories that start near the origin, whereas the stronger condition of asymptotic stability guarantees that trajectories converge to the origin when initialized close enough to the origin. 
The full set of initial conditions \( \xvec_0 \) leading to system trajectories that converge to the origin is called the \textit{stability region}, denoted \( \mathcal{D}_0 \subseteq \R^n \)~\cite{Haddad2008-hg}:
\begin{equation}\label{eqn:doa}
    \mathcal{D}_0 \triangleq \left \{ \xvec_0\in\R^n: \text{ if } \xvec(0) = \xvec_0, \text{ then } \lim_{t\rightarrow\infty}\xvec(t) = 0 \right \}.
\end{equation}
The conditions for stability in \Cref{def:stability} and the definition of the stability region in~\cref{eqn:doa} are given in terms of the long-term behavior of solutions to~\cref{eqn:nonlinear-sys} and are impractical to verify for general nonlinear dynamical systems. 
Instead, the following theorem of Lyapunov~\cite{lyapunov1992} provides sufficient conditions for stability and an estimate of the stability region in terms of a Lyapunov function \( V(x) \):
\begin{theorem}{(Lyapunov's Theorem~\cite{lyapunov1992})}\label{thm:lyapunov-thm}
    Consider the nonlinear dynamical system (\ref{eqn:nonlinear-sys}). If there exists a continuously differentiable function \( V:\R^n\to\R \) satisfying \( V(0)=0 \) and
    \begin{align*}
        V(\xvec) > 0, \quad \dot V(\xvec)=\nabla {V(\xvec)}^\top f(\xvec) < 0,
    \end{align*}
    for all \( \xvec \) in a neighborhood of the origin, then
    the origin is locally asymptotically stable. Furthermore, for any \( c>0 \), the set
    \begin{equation}
        \mathcal{D}(c) \triangleq \{\xvec: V(\xvec) \leq c, ~\dot{V} < 0\}
    \end{equation}
    is a subset of the stability region, i.e., \( \mathcal{D}(c) \subset \mathcal{D}_0 \).
\end{theorem}
The identification of \textit{any} Lyapunov function satisfying the conditions of \Cref{thm:lyapunov-thm} is a sufficient condition for system stability. Moreover, a subset of the true stability region can be characterized by a sublevel set of \( V \), as depicted in~\Cref{fig:doa-diagram}.
Zubov's theorem, introduced in the next section, provides conditions characterizing a \textit{specific} Lyapunov function that characterizes the entire stability region of the nonlinear system.

\subsection{Stability analysis based on Zubov's Theorem}\label{sec:2.3-zubov}
Zubov's theorem defines a specific Lyapunov function as the solution to a first-order PDE\@:
\begin{theorem}{(Zubov's Theorem~\cite{zubov1964,Haddad2008-hg})}\label{thm:zubov-theorem}
    Let \( h:\R^n\to\R \) be continuous with \( h(0)=0 \) and \( h(\mathbf{x})>0 \) for all \( \mathbf{x}\in\R^n\setminus \{ 0 \} \). Suppose there exists a continuously differentiable function \( V:\R^n\to\R \) satisfying \( V(0)=0 \) and 
    \begin{equation}\label{eqn:zubov-equation}
        \dot{V}(x) = \nabla {V(\xvec)}^\top f(\xvec) = -h(\xvec)[1-V(\xvec)].
    \end{equation}
    Then, if \( \mathcal{D}\subset\R^n \) contains the origin and
    \begin{enumerate}
        \item \( 0<V(\xvec)<1 \), for all \( x\in\mathcal{D} \), and 
        \item \( V(\xvec)\to 1 \) as \( \xvec\to\partial\mathcal{D} \) or \( \|\xvec \|\to\infty \),
    \end{enumerate}
    then the system (\ref{eqn:nonlinear-sys}) is asymptotically stable about the origin with the stability region of \( \mathcal{D}_0 = \mathcal{D} \).
\end{theorem}

The first-order PDE~\eqref{eqn:zubov-equation} is known as the \textit{Zubov equation}. 
One way to characterize the entire stability region of the nonlinear system is therefore to identify a Lyapunov function satisfying the Zubov equation and an associated domain \( \mathcal{D} \) satisfying the additional conditions in \Cref{thm:zubov-theorem}.
However, closed-form solutions to Zubov's equation have been found only for a limited number of systems.

For nonlinear systems with analytic \( f \), Zubov demonstrated that if \( h \) is a quadratic function, then the solution to the Zubov equation (\ref{eqn:zubov-equation}) can be represented as an infinite series:
\begin{equation}\label{eqn:zubov-inf-sum}
    V(\mathbf{x}) = V_2 + V_3 + \cdots = \sum_{i=2}^\infty V_i(\mathbf{x}),
\end{equation}
where \( V_i(\mathbf{x}) \) are homogeneous polynomial functions of \( \xvec \) with degree \( i \)~\cite{Hafstein2015,Hahn1967-it}. 
Truncating this series at degree \( p \) yields an approximation of the Lyapunov function:
\begin{equation}\label{eqn:zubov-trunc-sum}
    \widetilde{V}(\mathbf{x}) = V_2 + V_3 + \cdots + V_p = \sum_{i=2}^p V_i(\mathbf{x})
\end{equation}
which is a Lyapunov function satisfying the conditions of \Cref{thm:lyapunov-thm} and thus characterizes a subset of the stability region~\cite{zubov1964}. One way to compute a Lyapunov function is to use knowledge of \( f \) to recursively compute terms in this series~\cite{margolis1963}. In contrast, we propose a new method that truncates the series at \( p=2 \), yielding a quadratic form for the Lyapunov function, and we will infer the unknown quadratic coefficients from data without requiring knowledge of the system governing equations. 
\section{Physics-Informed Machine Learning for Characterizing the System Stability Region}\label{sec:3-lyapinf}
This section presents our physics-informed method for learning the stability region using our proposed Lyapunov function inference method (LyapInf) which infers a quadratic Lyapunov function from system trajectory data and provides us with an ellipsoidal estimate of the stability region. \Cref{sec:3.1:infer-qlf} introduces the mathematical formulation of the method. \Cref{subsec: implementation,sec:3.3-considerations} discuss implementation of the method and its dependence on user-defined inputs. \Cref{sec:3.4-doa-estimation} describes a Monte Carlo procedure for estimating the largest stability region associated with the inferred quadratic Lyapunov function.

\subsection{Mathematical formulation of the method}\label{sec:3.1:infer-qlf}
We assume that we have a data set of \( N \) pairs of state and time derivative data, \( {\{\xvec(t_i),\dot\xvec(t_i)\}}_{i=1}^N \), where the time derivative data may be extracted directly from a black box solver that evaluates \( f \) at \( \xvec(t_i) \), or approximated from the state data using standard finite difference schemes, e.g., 
\begin{equation}
    \dot{\mathbf{x}}(t_i) = \frac{\mathbf{x}(t_i) - \mathbf{x}(t_{i-1})}{t_i-t_{i-1}}, \quad \text{for } i = 1, 2, \ldots, N.
\end{equation}
To simplify the exposition, we restrict our notation in this section to the setting where the \( N \) data pairs arise from \( N \) timesteps of a single state trajectory. However, the formulation can be straightforwardly extended to data sets with multiple trajectories starting from different initial conditions. 

We seek to fit a quadratic Lyapunov function that minimizes the residual of the Zubov equation over the data. Or in other words, we inform physics knowledge from the system control domain in the form of the Zubov equation into a data-fitting problem. We assume \( f \) in~\eqref{eqn:nonlinear-sys} is analytic, and we define \( h(\xvec) = \xvec^\top\Qmat\xvec \), where \( \Qmat\in\R^{n\times n} \) is symmetric positive definite. We then adopt a quadratic form for an approximate solution to Zubov's equation, \( \widetilde{V}(\xvec) = \xvec^\top\Pmat\xvec \), where \( \Pmat \in \R^{n\times n} \) is also symmetric positive definite. Substituting the quadratic forms for \( h \) and \( \widetilde{V} \) into (\ref{eqn:zubov-equation}) and re-arranging terms yields the residual of the Zubov equation for our approximation:
\begin{equation}\label{eqn:zubov-residual}
    (\text{Residual})\quad \mathscr{R}(\Pmat,\xvec,\dot\xvec) = \dot{\xvec}^\top\Pmat\xvec + \xvec^\top\Pmat\dot{\xvec} + \xvec^\top\Qmat\xvec - \xvec^\top\Qmat\xvec\xvec^\top\Pmat\xvec.
\end{equation}
To fit \( \Pmat \) to data, we then solve
\begin{align}\label{eq: first residual minimization}
    \widetilde{\Pmat} = \arg\min_{\Pmat \in \mathcal{M}_n} \frac{1}{N} \sum_{i=1}^N\mathscr{R}(\Pmat,\xvec(t_i),\dot\xvec(t_i)),
\end{align}
where \( \mathcal{M}_n \) is the manifold of symmetric positive definite matrices in \( \R^{n\times n} \). 
The inferred Lyapunov function is then given by \( \widetilde V(\xvec) = \xvec^\top\widetilde{\Pmat}\xvec \). 

\subsection{Algorithmic implementation}\label{subsec: implementation}
We now describe how to implement the mathematical formulation of the minimization problem  from~\Cref{sec:3.1:infer-qlf} as a concrete algorithm. Let \( \otimes \) denote the Kronecker product, \( \mathrm{vec}(\cdot):\R^{m\times n}\to\R^{mn} \) the vectorization (\emph{flattening}) of a matrix, and \( \mathrm{mat}(\cdot):\R^{mn}\to\R^{m\times n} \) the inverse operator of vectorization. For computational implementation of the mathematical procedure we have described, we will find it convenient to define \( \pvec = \mathrm{vec}(\Pmat) \) and \( \qvec = \mathrm{vec}(\Qmat) \), as well as the \( k \)-times repeated Kronecker product of the state vector:
\begin{equation}
    \xvec^{\otimes k} = \underbrace{\xvec \otimes \cdots \otimes \xvec}_{k-\text{times}} \in \R^{n^k},
\end{equation}
e.g., \( \xvec \otimes \xvec = \xvec^{\otimes 2} \). Using Kronecker product identities from~\cite{brewerKroneckerProductsMatrix1978}, we can rewrite \cref{eqn:zubov-residual} as:
\begin{equation}\label{eqn:zubov-residual-2}
    \mathscr{R}(\Pmat,\xvec,\dot\xvec) = \pvec^\top\xvec^{\circledast} + \qvec^\top\xvec^{\otimes 2}  - {(\qvec\otimes\pvec)}^\top\xvec^{\otimes 4}
\end{equation}
where \( \xvec^{\circledast} = \dot{\xvec}\otimes\xvec+\xvec\otimes\dot{\xvec} \). 
This lets us reformulate~\eqref{eq: first residual minimization} as the following constrained minimization:
\begin{equation}
    \begin{gathered}
        \widetilde{\pvec} = \arg\min_{\pvec\in\R^{n^2}}\frac{1}{N}\sum_{j=1}^N \left \| \pvec^\top\xvec_j^{\circledast} + \qvec^\top\xvec^{\otimes 2}_j  - {(\qvec\otimes\pvec)}^\top\xvec^{\otimes 4}_j \right \|_2^2 \quad  
        \text{subject to } \Pmat = \mathrm{mat}(\pvec) = \Pmat^\top > 0.
    \end{gathered}    
\end{equation}
This reformulation makes clear that the optimization objective and constraints are both convex, enabling us to compute the solution of the constrained optimization using existing convex optimization packages. 
In order to use such packages, we find it convenient to reformulate the optimization in terms of the Frobenius norm of the matrix whose columns contain the residuals over all data. The optimization problem can then be written as:
\begin{equation}\label{eqn:nonintrusive-lyapinf}
    \begin{gathered}
        \widetilde{\pvec} = \arg\min_{\pvec\in\R^{n^2}} \frac{1}{N} \left \| \pvec^\top\Xmat^{\circledast} + \qvec^\top\Xmat^{\otimes 2} - {(\qvec\otimes\pvec)}^\top\Xmat^{\otimes 4} \right \|_F^2 \quad
        \text{subject to } \Pmat = \mathrm{mat}(\pvec) = \Pmat^\top > 0
    \end{gathered}
\end{equation} 
where we define the data matrices
\begin{gather}
    \Xmat^{\circledast} = \begin{bmatrix}
        | & & | \\
        \xvec^{\circledast}(t_1) & \cdots & \xvec^{\circledast}(t_N)\\
        | & & |
    \end{bmatrix} \in \R^{n^2 \times N}
    \\[0.5em]
    \Xmat^{\otimes k} = \begin{bmatrix}
        | & & | \\
        \xvec^{\otimes k}(t_1) & \cdots & \xvec^{\otimes k}(t_N) \\
        | & & |
    \end{bmatrix} \in \R^{n^k \times N}.
\end{gather}

In our numerical experiments in~\Cref{sec:4-examples}, we solve~\eqref{eqn:nonintrusive-lyapinf} in the Julia language using the JuMP~\cite{Lubin2023} and Splitting Conic Solver~\cite{ocpb:16} packages. To enforce positive semidefiniteness, we impose the convex constraint \( \Pmat \succeq 0 \). We then promote strict positive definiteness by additionally requiring the diagonal elements of \(\Pmat \) to satisfy \( P_{ii} \;\ge \; \varepsilon \) for all \( i \) where \(\varepsilon>0 \) is a small constant. Although a diagonal lower bound by itself guarantees only a positive trace, when combined with \(\Pmat\succeq0\), it effectively pushes all eigenvalues away from zero. Exploring alternative implementation approaches, including manifold optimization or Cholesky factor parameterizations of \( \Pmat \), is an area for future work.

\subsection{Algorithm input considerations}\label{sec:3.3-considerations}
Both the training data and the choice of auxiliary function \( h \) are algorithm inputs which influence the performance of the method. This section discusses these considerations and describes the choices used in our numerical experiments.

\subsubsection{Training Data}
The inferred Lyapunov function and associated estimate of the stability region will vary depending on the training data set. For the numerical experiments in this work, we follow an approach consistent with recent data-driven methods for estimating stability regions based on Zubov's theory~\cite{meng2023Learning,liu2025physics,gruneComputingLyapunovFunctions2021}, which collect data by randomly sampling from a specified operating regime \( \mathcal{X}\subset\R^n \). To generate training data for LyapInf, we simulate the system for initial conditions \( \xvec_0\in\mathcal{X} \) and collect the trajectory data at each timestep. We then discard the points of the trajectory outside the region of interest \( \mathcal{X} \). Developing further analytical understanding of our method's dependence on training data, and potentially new targeted data acquisition strategies, is an area for future work. 

\subsubsection{Choice of Auxiliary Function \texorpdfstring{\( h \)}{}}
The choice of the quadratic operator \( \Qmat \) defining the auxiliary function \( h \) will also influence the resulting inferred Lyapunov function and associated stability region. In this work, we assume \( \Qmat = \gamma \Id_n \), where \( \gamma > 0 \). This scalar \( \gamma \) modulates the influence of the quartic term \( -\xvec^\top\Qmat\xvec\xvec^\top\Pmat\xvec \) in \cref{eqn:nonintrusive-lyapinf}, ensuring that the quartic scaling does not disproportionately affect the objective function during optimization. We determine \( \gamma \) by testing values within [1e-3, 10] and selecting the \( \gamma \) giving the largest estimated stability region. Further theoretical and numerical exploration of the effect of the choice of \( \Qmat \) on our method is a direction for future work.

\subsection{Monte Carlo Estimation of the Stability Region}\label{sec:3.4-doa-estimation}
As discussed in \Cref{sec:2.2-lyapunov-stability-theory,sec:2.3-zubov}, the set \( \mathcal{D}(c)= \{x: \widetilde{V}(\xvec) \leq c, \dot{\widetilde{V}}<0\} \) associated with our inferred quadratic Lyapunov function \( \widetilde V(\xvec) \) is a subset of the stability region.
Because \( \widetilde{V} = \xvec^\top\Pmat\xvec \) with positive definite \( \Pmat \), the sublevel set \( \{x: \widetilde V(x)\leq c\} \) is an ellipsoid which expands as \( c \) increases.
To provide the largest possible ellipsoid estimate of the stability region, we seek the maximum \( c \) such that the sublevel ellipsoid remains inside the \( \dot{\widetilde{V}}<0 \) region. This maximal \( c \), denoted \( c_* \) is the solution to the following optimization problem~\cite{Hachicho2007,Matallana2010}:
\begin{equation}\label{eqn:soa-minimization-problem}
    c_* = \min_{\xvec \in \R^n} \widetilde{V}(\xvec) \quad \text{subject to } \dot{\widetilde{V}}(\xvec) = 0.
\end{equation}
Unfortunately, this optimization is non-convex and generally challenging to globally solve~\cite{Hachicho2007}. Instead, for the relatively low-dimensional examples considered here, we use the Monte-Carlo sampling method proposed by~\cite{Najafi2016} to estimate \( c_* \). This brute-force approach uniformly samples points within a region of interest and sets \( c_* \) to be the largest value such that all sampled points within the sublevel set satisfy \( \dot{\widetilde{V}}<0 \). 
One drawback of this approach is that it can require many samples to provide a good \( c_* \) estimate, particularly for higher-dimensional systems. 
\section{Numerical Experiments}\label{sec:4-examples}
We apply our proposed method to a series of benchmark nonlinear dynamical systems, described in~\Cref{sec:4.1-systems}. \Cref{sec:4.3-discussion} presents and discusses the results.

\subsection{Test Problems}\label{sec:4.1-systems}
This section describes six nonlinear dynamical systems commonly used as benchmarks for nonlinear stability analysis that serve as test problems for our method.  After introducing each system's governing equations, we describe the initial conditions used to generate system state trajectories used as training data for the method. For each test problem, \Cref{tab:problem-parameters} tabulates the timestep size and final time for the trajectory data, as well as the region of interest \( \mathcal{X} \), and the scaling \( \gamma \) that defines the auxiliary quadratic operator \( \mathbf{Q} \). Trajectory data are obtained by numerical integration using a fourth-order Runge-Kutta method, and time derivative data are directly extracted from the solver. 

\subsubsection{2D Quadratic System} This first example from~\cite{Tesi1996,Hachicho2007,Najafi2016} is defined by
\begin{equation}
    f(\xvec) = \begin{bmatrix}
        -2x_1 + x_1x_2  \\
        -x_2 + x_1x_2 
    \end{bmatrix} .
\end{equation}
The training data are generated using \( M=16 \) different initial conditions, equally spaced along a circle of radius 5 centered at the origin and all lying within the domain \( \mathcal{X} = [-5,5]^2\).

\subsubsection{Van der Pol Oscillator}
\begin{equation}
    f(\xvec) = \begin{bmatrix}
        -x_2 \\ 
        x_1 - \mu(1-x_1^2)x_2
    \end{bmatrix}
\end{equation}
where \( \mu=1 \) is the damping parameter. The training data are generated using \( M=10 \) different initial conditions, equally spaced along a circle of radius 1.5 centered at the origin and all lying within the domain \( \mathcal{X} = [-3,3]^2 \).

\subsubsection{Nonlinear Pendulum}
\noindent
\begin{equation}
    f(\xvec) = \begin{bmatrix}
        x_2 \\
        - \sin(x_1) - 0.5x_2
    \end{bmatrix}.
\end{equation}
The training data are generated using \( M=20 \) different initial conditions, equally spaced along the boundary of the domain \( \mathcal{X} = [-4,4]^2 \).

\subsubsection{Trig-Exp Nonlinear System}
The fourth example includes high nonlinearity with cosine and exponential terms. This system is an example from~\cite{Najafi2016,chesi2009estimating,Matallana2010} defined by
\begin{equation}
    f(\xvec) = \begin{bmatrix}
        -x_1+x_2 + 0.5(\exp(x_1)-1) \\
        -x_1-x_2 + x_1x_2 + x_1\cos(x_1) 
    \end{bmatrix}.
\end{equation}
The training data are generated using \( M=30 \) different initial conditions uniformly sampled from the domain \( \mathcal{X} = [-3,3]^2 \).

\subsubsection{3D Cubic System}
This example is a 3D system with cubic nonlinearity from~\cite{Hachicho2007,Najafi2016} defined by
\begin{equation}
    f(\xvec) = \begin{bmatrix}
        -x_1 + x_2x_3^2  \\
        -x_2 - x_1x_2 \\
        -x_3 
    \end{bmatrix} .
\end{equation}
The training data are generated using \( M=25 \) different initial conditions, equally spaced along the surface of a sphere with radius 3 centered at the origin and all lying within the domain \( \mathcal{X} = [-3,3]^3 \).
    
\begin{table}[t]
    \centering
    \caption{Test problem parameters.}
    \begin{tabular}{lcccc}\toprule
        {System} & {\( \mathcal{X} \)} & {\( t_f \)} & {\( \Delta t \)} & {\( \gamma \)} \\ \midrule
        2D Quadratic                            & \( {[-5,5]}^2 \)    & \( 5 \)  & 0.01  & 1    \\[0.2em]
        Van der Pol                             & \( {[-3,3]}^2 \)    & \( 5 \)  & 0.01  & 2.0  \\[0.2em]
        Nonlinear Pendulum                      & \( {[-4,4]}^2 \)    & \( 10 \) & 0.001 & 0.2  \\[0.2em]
        Trig-Exp Nonlinear                      & \( {[-3,3]}^2 \)    & \( 10 \) & 0.01  & 0.01 \\[0.2em]
        3D Cubic                                & \( {[-3,3]}^3 \)    & \( 5 \)  & 0.01  & 1    \\[0.2em]
        Networked Van der Pol                   & \( {[-4,4]}^{20} \) & \( 10 \) & 0.01  & 0.1  \\ \bottomrule
    \end{tabular}\label{tab:problem-parameters}
\end{table}

\subsubsection{20D Networked Van der Pol Oscillator}
Inspired by~\cite{liu2025physics} and~\cite{kundusos2015}, this example consists of 10 Van der Pol oscillator subsystems \( {\{N_i\}}_{i=1}^{10} \), where each is defined by:
\begin{equation}
    f_i(\xvec) = \begin{bmatrix}
        -x_{i2} \\
        x_{i1} - \mu_i(1-x_{i1}^2)x_{i2} + \displaystyle{\sum_{j\neq i}}\zeta_{ij}x_{i1}x_{j2}
    \end{bmatrix}
\end{equation}
where \( \mu_i \) is the damping parameter uniformly sampled within the parameter domain \( [0.5,2.5] \) for each subsystem. The interconnection strengths between subsystems \( i \) and \( j \) are defined by parameters \( \zeta_{ij} \) that are set to 0 with a 50\% probability, indicating no connection, or, with a 50\% probability, are uniformly chosen from \( [-0.1,0.1] \). The data are generated using \( M=80 \) different initial conditions sampled within the domain \(\mathcal{X}=[-4,4]^{20} \). Specifically, for each subsystem we choose points along a circle of radius one.

As mentioned in~\Cref{sec:3.4-doa-estimation}, the Monte-Carlo procedure to estimate \( c_* \) is a challenge for high-dimensional systems. To estimate \( c_* \) for our 20D networked Van der Pol oscillator example, we therefore apply this Monte Carlo algorithm to estimate \( c_* \) within the 10 individual two-dimensional state spaces associated with each subsystem while fixing the other subsystem coordinates at zero. We then report the results of the most conservative \( c_* \) over all 10 subsystems. 
Future work will investigate alternative strategies for estimating the stability region associated with our inferred Lyapunov function that scale to higher dimensions. Finally, we note that the method implementation as well as numerical examples are available online\footnote{Github repository: \href{https://github.com/smallpondtom/LyapunovFunctionInference.jl}{https://github.com/smallpondtom/LyapunovFunctionInference.jl}}.

\subsection{Results and Discussion}\label{sec:4.3-discussion}

\begin{figure}[!htb]
    \centering
    \includegraphics[width=0.9\columnwidth]{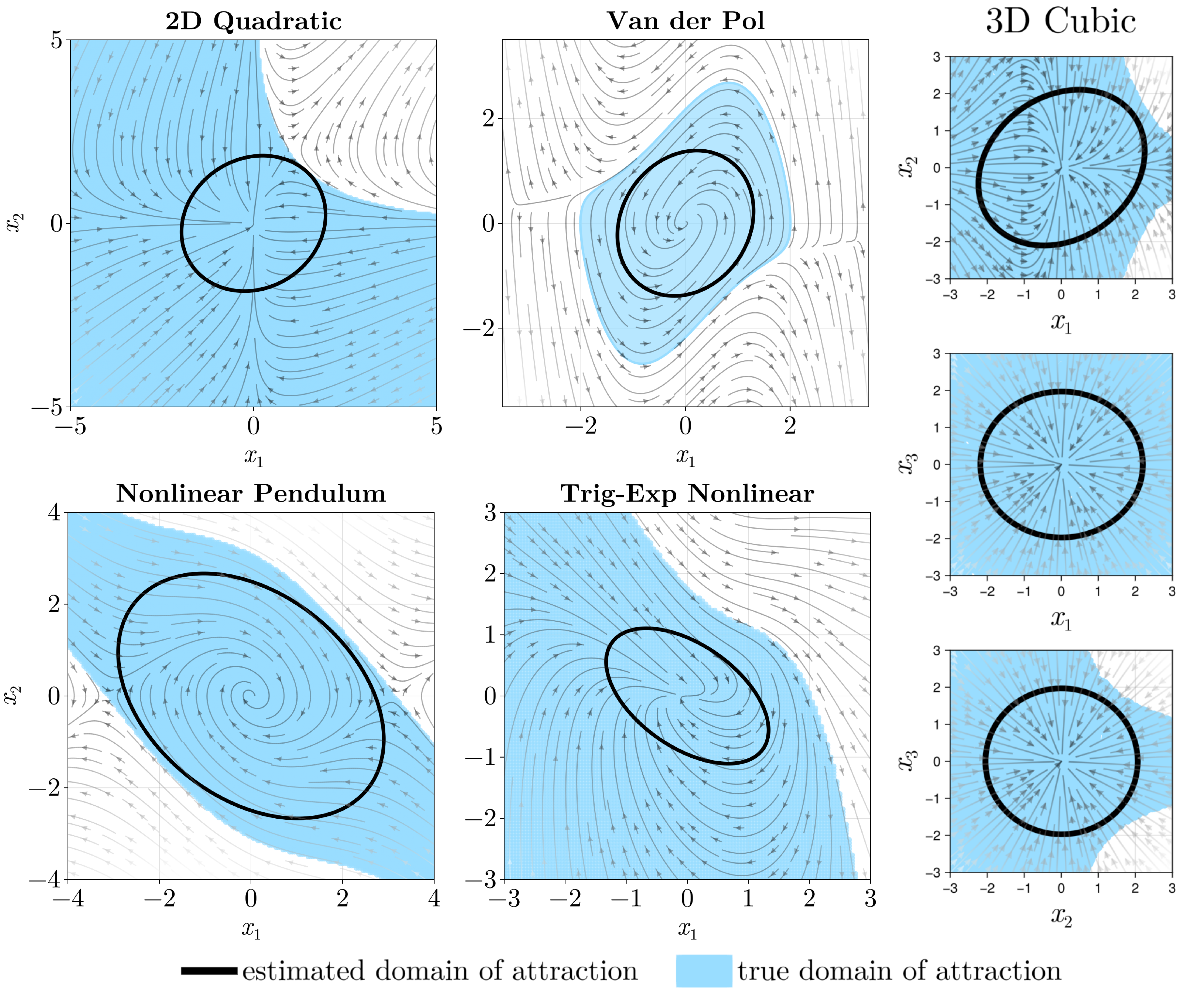}
    \caption{Overlay of the estimated stability region (interior of black ellipse) and true stability region for 2D numerical examples, and 2D cross-sections of each axis plane for the 3D example.}\label{fig:numerical-doa}
\end{figure}

\begin{figure}[!htb]
    \centering
    \includegraphics[width=0.8\columnwidth]{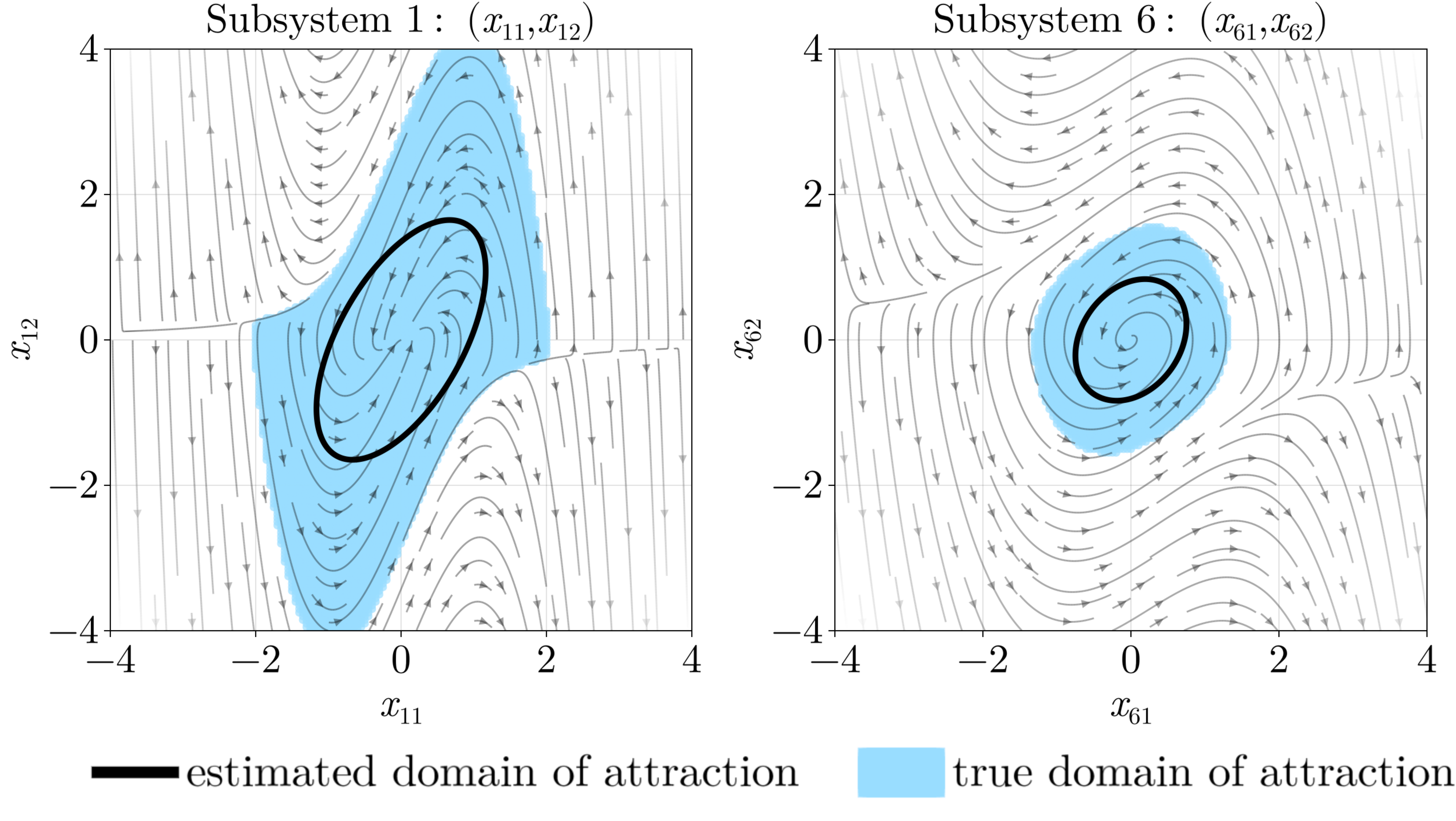}
    \caption{Estimated stability region (interior of black ellipse) for subsystems 1 and 6 of the 20D networked Van der Pol oscillator.}\label{fig:networked-vanderpol}
\end{figure}

\Cref{fig:numerical-doa} plots the estimated stability region \( \mathcal{D}(c_*) \) obtained using the Monte Carlo procedure described in \Cref{sec:3.4-doa-estimation} for the two- and three-dimensional test problems (Examples 1--5): the estimated stability region is the interior of the black ellipse in each plot. For comparison, the shaded blue region in each plot is the true stability region \( \mathcal{D}_0 \) obtained by direct numerical evaluation for these simple examples. 
For the 3D cubic system (\Cref{fig:numerical-doa}, bottom panel) and the 20D networked Van der Pol system, (\Cref{fig:networked-vanderpol}) we plot the projection of the true and estimated stability regions onto axial planes. For the Van der Pol system, we select the axial planes associated with two-dimensional subsystems and show only results for the first and sixth subsystems, which are representative of the remaining subsystems.
Our estimated stability regions are fully contained within the true stability regions. While the estimated stability regions are conservative, they are not overly so: the edges of our estimates are near the boundaries of the true stability region. 
We note that these ellipsoid estimates are, however, more conservative than estimates obtained in~\cite{kang2024data,liu2025physics}, which use neural networks to approximate Lyapunov functions in Zubov-based approaches. 
This is unsurprising given the greater expressive power of neural networks compared to our quadratic ansatz; the trade-off is the greater data requirements for training neural networks. For example,~\cite{liu2025physics} reports using 300,000 data to train the neural network Lyapunov function for the Van der Pol and networked Van der Pol systems, whereas we can infer a quadratic Lyapunov function using just 5,010 and 80,080 data for these systems, respectively. 
\section{Conclusion}\label{sec:conclusion}
We have presented a new physics-informed machine learning method for characterizing the stability region of nonlinear dynamical systems to enable the design of systems with known safe operation regimes.
The method learns a stability region by inferring a Lyapunov function from state trajectory data, and it is based on embedding Zubov's theorem, which characterizes a Lyapunov function as the solution to a first-order PDE known as the Zubov equation, into the training process.
The proposed method adopts a quadratic approximation for the solution to the Zubov equation and minimizes the average residual of the equation over the data to infer the unknown quadratic operator. 
We then learn an elliptical stability region characterized by the inferred quadratic Lyapunov function, which is an estimate of the true stability region, indicating a safe region of operation for a given physical system.
The method does not require explicit knowledge of the system's governing equations to estimate the stability region; instead, it leverages physics-informed machine learning to discover operators of Lyapunov functions from data or the underlying physics, making it applicable to black-box systems.
Numerical experiments on benchmark examples demonstrate that the method learns near-maximal ellipsoidal estimates of the true stability region for the tested systems with associated inferred quadratic Lyapunov functions.

There are several directions that are the subject of ongoing and future work.
We note that like many physics-informed machine learning methods, the performance of our physics-informed stability analysis method is sensitive to both the quality and quantity of data.
In our experiments, reducing the number of trajectories in the training data set led to poor estimates of the stability region, likely due to the data covering less of the system phase space.
On the other hand, the inclusion in the data set of many unstable trajectories can also potentially lead to poor results, although our experiments do use training data containing some unstable trajectories.
Further development of efficient strategies for collecting appropriate training data is a direction of active investigation.

Other research directions include developing computational procedures for efficiently learning higher-order polynomial Lyapunov functions to characterize stability regions with complex geometries, and efficiently estimating these regions for high-dimensional systems arising in multidisciplinary engineering design.
For high-dimensional systems such as those arising from PDE discretization, our framework could be applied to projection-based reduced models~\cite{Kramer2021,peherstorferDatadrivenOperatorInference2016,berkoozProperOrthogonalDecomposition1993,bennerSurveyProjectionBasedModel2015a} to analyze their stability properties or inform stability guarantees~\cite{kaptanoglu2021promoting,sawant2023Physicsinformed,goyal2025guaranteed,gkimisis2025Nonintrusive,peng2025extending}.
\section*{Acknowledgments}
This work was supported by the US Department of Energy Office of Science Energy Earthshot Initiative as part of the ‘Learning reduced models under extreme data conditions for design and rapid decision-making in complex systems’ project under award number DE-SC0024721.

\bibliography{refs}

\end{document}